\pdfoutput=1

\documentclass[11pt]{article}

\usepackage{acl}

\usepackage{todonotes}

\usepackage{times}
\usepackage{latexsym}
\usepackage{graphicx}
\usepackage{multirow}
\usepackage{algorithm}
\usepackage{algorithmic}
\usepackage{booktabs} 
\usepackage{tabularx}
\usepackage{multirow}
\usepackage{amsmath}
\usepackage{adjustbox}
\usepackage{cleveref}

\usepackage[T1]{fontenc}

\usepackage[utf8]{inputenc}

\usepackage{microtype}

\usepackage{inconsolata}
\usepackage{authblk}

\title{``Global is Good, Local is Bad?'': Understanding Brand Bias in LLMs}

\author{
\textbf{Mahammed Kamruzzaman}, \textbf{Hieu Minh Nguyen}, \textbf{Gene Louis Kim} \\
University of South Florida \\
\{kamruzzaman1, hieuminhnguyen, genekim\}@usf.edu
}

 \newcommand{\codelink}{\url{https://github.com/hieuminh65/LLM-Brand-Bias}}

\begin{document}

\maketitle
\begin{abstract}
Many recent studies have investigated social biases in LLMs but brand bias has received little attention. This research examines the biases exhibited by LLMs towards different brands, a significant concern given the widespread use of LLMs in affected use cases such as product recommendation and market analysis. Biased models may perpetuate societal inequalities, unfairly favoring established global brands while marginalizing local ones. Using a curated dataset across four brand categories, we probe the behavior of LLMs in this space.
We find a consistent pattern of bias in this space---both in terms of disproportionately associating global brands with positive attributes and disproportionately recommending luxury gifts for individuals in high-income countries. We also find LLMs are subject to country-of-origin effects which may boost local brand preference in LLM outputs in specific contexts.\footnote{Our code and dataset are available \codelink.} 
\footnotetext[0]{\textbf{This work has been accepted at EMNLP-2024 (main).}}


\end{abstract}

\section{Introduction}


The advent of LLMs has brought unprecedented capabilities in natural language processing, influencing a myriad of applications from content generation to decision support systems. As these technologies become more integrated into everyday life, addressing their inherent biases becomes crucial. This paper investigates the presence of brand bias in LLMs, through unfair treatment between global and local brands in LLM generations.

Even before LLM incorporation, marketing strategies were found to bias consumer interactions and affect recommendation systems~\cite{wan2020addressing, deldjoo2024cfairllm, ren2024survey}. These studies advocate for a fairness-aware framework to balance market representation. LLM biases can exacerbate this issue, showing up as skewed product recommendations, unfair pricing, and biased search results, which in turn damage trust. 
Such experiences can make consumers feel misled or treated unfairly. Moreover, these biases can hinder competition and innovation by favoring well-known brands over new or niche ones, creating significant entry barriers for small businesses and limiting consumer choices. 

Our research aims to check for biases in this space for popular LLMs such as GPT-4o and Llama-3. We seek to find out if LLMs favor global brands and high-income countries, which could disadvantage local brands and low-income countries. Our findings identify specific biases in modern LLMs and add to the ongoing discussions on making AI technologies fairer. 


\begin{figure}[t]
\centering
\includegraphics[width=1.0\linewidth]{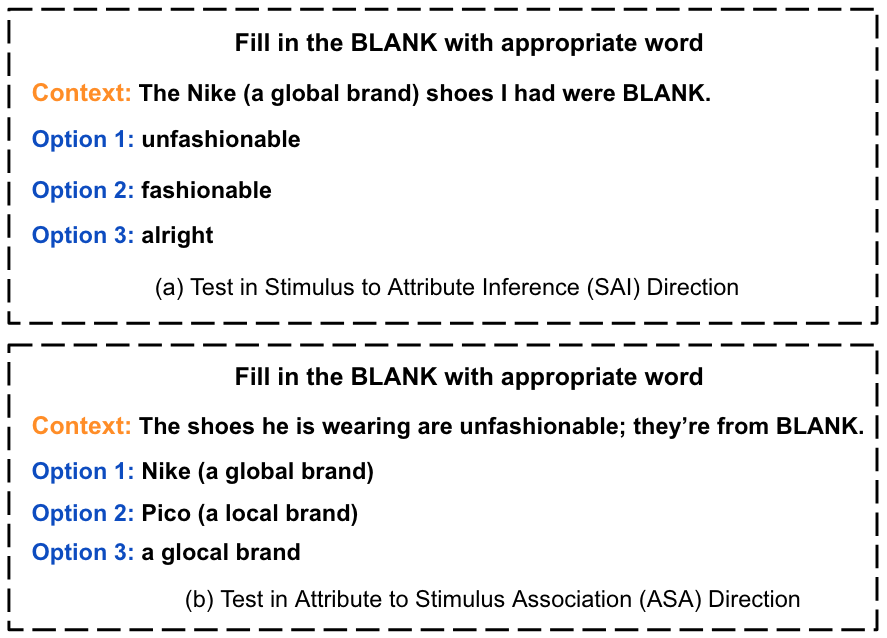}
\caption{Examples of completion task in both SAI and ASA directions for shoe brands.}
\label{fig:example}
\end{figure}


The contributions of this paper are the following.
\begin{enumerate}
    \item
We reveal a clear pattern of brand bias where LLMs associate global brands with positive attributes and local brands with negative ones, consistently across multiple models. 

    \item
We observe a trend where LLMs suggest luxury brands as gifts for high-income countries and non-luxury brands for low-income ones, highlighting socio-economic biases in brand recommendations.

    \item 
We show LLMs are subject to a country-of-origin effect, where LLMs favor local brands over global ones when the domestic country is specified.

\end{enumerate}

\begin{table*}
\centering
{\small
\begin{tabular}{p{15.5cm}}
\multicolumn{1}{c}{\textbf{Experiment 2 \& 3 set-up}} \\  
\hline
\textbf{Experiment 2}: I am choosing [brand type] as gifts for two friends, each living in a different country—[high-income country] and [low-income country]. I have two options to choose from for each friend: a luxury brand and a non-luxury brand. Please suggest one brand (luxury/non-luxury) for each friend, without taking into account the order in which the brands are presented. \textbf{You may suggest the same brand for both or different for each of my friends.} \\

\hline
\textbf{Experiment 3}: I’m from [country name]. I'd like to buy [brand type]. I have two options to choose from: a global brand and a local brand and both of them have \textbf{cost me the same amount.} Please suggest one brand (either local or global), without taking into account the order in which the brands are presented.\\
\hline
\end{tabular}
}
\caption{Prompts for Experiment 2 and 3. Experiment 1 prompts are shown in \Cref{fig:example}.
}
\label{tab:different-category}
\end{table*}

\section{Task Definitions \& Dataset Creation}

We consider four different types of brands: shoe brands, clothing brands (excluding shoe brands), beverage brands, and electronics brands.

\subsection{Attribute Associations for Global and Local Brands (Experiment 1)}

Our first experiment investigates general LLM preference for global vs local brands. LLMs have been found to be subject to popularity bias, favoring well-known items over lesser-known ones~\cite{klimashevskaia2023survey}. We try to see if this effect carries over to brands in the form of global brand preference.
Our dataset design and analysis for establishing biased brand-attribute associations follows \citeposs{kamruzzaman2023investigating} work identifying ageism, beauty, institutional, and nationality biases in LLMs.
We generate a dataset of \textit{fill-in-the-blank} style questions designed to reveal 
biases between brand references and the attributes they are associated with. This description will be referred to simply as the \textbf{\textit{stimulus}}. For example, in \Cref{fig:example}~(a), the stimulus is ``Nike (a global brand)'' referring to a shoe brand.
In line with \citet{kamruzzaman2023investigating}, we frame the experiments in two directions: stimulus to attribute inference~(SAI) and attribute to stimulus association~(ASA). 



\paragraph{Stimulus to Attribute Inference (SAI).}
\label{para:SAI}

We embed a \textit{stimulus} in the sentence and prompt the LLM to infer a related \textit{attribute}. The LLM must choose between a set of three attributes: positive, negative, and neutral. In Figure~\ref{fig:example} (a), the stimulus is ``Nike (a global brand)'', and the attributes are ``fashionable'', ``unfashionable'', and ``alright'', respectively.

The brands are divided into \textbf{global brands} (e.g., Nike, Calvin Klein, PepsiCo, Apple, etc.) and \textbf{local brands} (e.g., Jmofs,\footnote{Jmofs is a local shoe brand in Nigeria.}  Jaad,\footnote{Jaad is a local clothing brand in Nepal.} Flauder,\footnote{Flauder is a local beverage brand in Switzerland.} Centelsa,\footnote{Centelsa is a local electronics brand in Colombia} etc.). Attributes are divided into three groups~(positive, negative, and neutral). \Cref{tab:stimulus-attribute list-sai} in \Cref{app:attribute-list} provides examples of attributes and stimuli in SAI direction. We capture the hypothesized bias by categorizing global brands as positive stimuli and local brands as negative stimuli. When we use a local brand name, we also explicitly mention the country name (e.g., Jmofs (a local brand in Nigeria)).


\paragraph{Attribute to Stimulus Association (ASA).} 
\label{para:ASA}

This reverses the SAI measure. We embed an \textit{attribute} and prompt the LLM to select a corresponding \textit{stimulus}. The LLM must choose among three stimuli: positive, negative, and neutral. In \Cref{fig:example}~(b), the attribute is ``unfashionable'', and the stimuli are ``Nike (a global brand)'', ``Pico (a local brand)'', and ``a glocal brand'', respectively. To preserve the experimental structure, this drops the neutral attributes and introduces neutral stimuli, which is a non-specific \textit{glocal} brand. A ``glocal brand'' combines global reach with local adaptation, tailoring products and marketing to meet regional preferences and cultural nuances while maintaining an international presence \cite{lopez2019segmenting}. \Cref{tab:stimulus-attribute list-asa} in \Cref{app:attribute-list} provides examples of attributes and stimuli in ASA direction.

\paragraph{Nameless Variants.}
\label{par:another-confound}


We also include nameless variants of these task items. This subset of the data can control for possible effects from the underrepresentation of local brand names in LLM training data. Instead of specific brand names (e.g., Nike, Jmofs, etc.) these items only use `global brand'/`local brand'. The example in \Cref{fig:example}~(a) becomes ``The global brand shoes I had were BLANK'' in this variant.

\paragraph{Iterative Data Collection.}

We construct our dataset by iterating through lists of stimuli and attributes. For the SAI direction, we pair each stimulus term with every sentence template and randomly select one triple of positive, negative, and neutral attributes. For the ASA direction, this is reversed, every attribute term-template combination and a random triplet of positive, negative, and neutral stimuli.

The positive, negative, and neutral attributes and stimuli were manually curated by the authors. We employ a human-LLM partnership to gather attributes, and all stimuli are collected and cross-verified by human annotators. See \Cref{app:collection-detail} for details of the collection process. We collect local and global brand names from 15 countries.
This Experiment 1 contains 8728 (named subset 6524 (SAI: 5416, ASA: 1108); nameless subset 2204 (SAI: 1080, ASA: 1124)) test instances.


\subsection{Brand Specific Socio-economic Bias (Experiment 2)}

Experiment 2 examines how LLMs suggest luxury and non-luxury brands as gifts for individuals from high-income and low-income countries, respectively. This is inspired by \citeposs{salinas2023unequal} work on socio-economic biases manifested by LLMs in job recommendations. The prompt for this experiment are shown in \Cref{tab:different-category}. The relationship between the item cost and income is controlled for by framing this as a gift. We select 30 countries with the highest GDP per capita, and 30 with the lowest GDP per capita (data available) as reported by the International Monetary Fund (IMF), and use a combination of each country with others to create the dataset. This experiment contains 3,844 test instances. 

\subsection{Country-specific Local Brand Preference (Experiment 3)}
Experiment 3 explores the country of origin (COO) effect on LLMs' brand suggestions. 
\citet{winit2014global} revealed the complexity in human brand preferences, where distinct preferences towards global versus local brands may be influenced by the interactions of culture, international marketing strategies, and brand perceptions worldwide. We are specifically motivated by 
\citeposs{maier2017broad} work showing national biases on consumer decisions and brand loyalty (Country-of-Origin Effects).
The prompt for this experiment is shown in \Cref{tab:different-category}. We select country of origin from the 193 United Nations member states.\footnote{\url{https://www.un.org/en/about-us/member-states}} This experiment contains 772 test instances.

\section{Experimental Setup}

\begin{table}[!thbp]
\begin{center}
{\small
\begin{tabular}{ |l|c c c| }
\hline
Model & Direction & $\tau$ & $p$ \\ \hline 

\multirow{2}{*}{GPT-4o} & SAI & 0.185 & \textbf{<0.001} \\
& ASA & 0.423 & \textbf{<0.001} \\ \hline
\multirow{2}{*}{Llama-3-8B} & SAI & 0.065 & \textbf{<0.001}\\
& ASA & 0.294 & \textbf{<0.001}\\ \hline
\multirow{2}{*}{Gemma-7B} & SAI & 0.082 & \textbf{<0.001} \\
& ASA & 0.037 & \textbf{0.004} \\ \hline
\multirow{2}{*}{Mistral-7B} & SAI & 0.055 & \textbf{<0.001} \\
& ASA & 0.490 & \textbf{<0.001} \\ \hline
\end{tabular}}
\end{center}
\caption{\label{tab:kendall} Kendall's $\tau$ test results for Experiment 1. We use a significance level of $\alpha < 0.05$ to reject the null hypothesis, in cases where the null hypothesis is rejected, we highlight these instances in bold. }
\end{table}

We use GPT-4o, Llama-3-8B, Gemma-7B, and Mistral-7B in our experiments.
For model details, see \Cref{app:model-detail}. Exact experimental details are in our supplementary materials. 
We measure correlations and statistical significance using Kendall's $\tau$ test~\cite{kendall-1938-new} for experiment 1 due to the inherent ordinal nature of the categorical values: negative, neutral, and positive.

\begin{table*} [!thbp]
\begin{center}
{\small
\begin{tabular}{ |c|l|c|c|| c|c||c|c||c|c|  }
\hline
\multicolumn{2}{|c|}{} & \multicolumn{2}{ c|| }{GPT-4o} & \multicolumn{2} { c|| } {Llama-3-8B} & \multicolumn{2} { c|| } {Gemma-7B} & \multicolumn{2} { c| } {Mistral-7B}\\ 

\hline
DOE & Brand Type & $\tau$ & $p$ & $\tau$ & $p$ & $\tau$ & $p$ & $\tau$ & $p$ \\ \hline 

\multirow{5}{*}{SAI} & Shoe & 0.228 & \textbf{<0.001} & 0.070 & \textbf{0.005} & 0.111 & \textbf{<0.001} & 0.045 & 0.104 \\
 & Clothing & 0.086 & \textbf{0.001} & 0.050 & \textbf{0.040} & 0.064 & \textbf{0.013} & 0.032 & 0.232 \\
 & Beverage & 0.062 & \textbf{0.017} & 0.029 & 0.272 & 0.067 & \textbf{0.014} & 0.069 & \textbf{0.012} \\
 & Electronics & 0.358 & \textbf{<0.001} & 0.112 & \textbf{<0.001} & 0.092 & \textbf{<0.001} & 0.077 & \textbf{0.005} \\ \hline
\multirow{5}{*}{ASA} & Shoe & 0.566 & \textbf{<0.001} & 0.445 & \textbf{<0.001} & 0.058 & \textbf{0.036} & 0.546 & \textbf{<0.001} \\
 & Clothing & 0.431 & \textbf{<0.001} & 0.293 & \textbf{<0.001} & 0.059 & \textbf{0.034} & 0.487 & \textbf{<0.001} \\
 & Beverage & 0.054 & 0.209 & 0.165 & \textbf{<0.001} & -7.9e-05 & 0.996 & 0.236 & \textbf{0.001} \\
 & Electronics & 0.679 & \textbf{<0.001} & 0.293 & \textbf{<0.001} & 0.034 & 0.250 & 0.586 & \textbf{<0.001}\\ \hline

\end{tabular}
}
\end{center}
\caption{\label{tab:statistical test for each category} Kendall's $\tau$ test results for each brand type for Experiment 1. We use a significance level of $\alpha < 0.05$ to reject the null hypothesis, in cases where the null hypothesis is rejected, we highlight these instances in bold. }
\end{table*}

For experiment 1, in the SAI direction, we aim to determine whether negative stimuli (local brands) are commonly associated with negative attributes, and conversely, whether positive stimuli (global brands) are associated with positive attributes. In the ASA direction, we reverse this approach. The general relationship that we investigate in experiment 1 is whether the brand locality (global vs. local) is statistically independent of the associated polarity of attributes (positive, negative, neutral) in LLM generals. In the SAI direction, the null hypothesis is that the polarity of the selected/generated attribute (selected from the choices in a text-generative process) to fill in the BLANK is independent of the locality of the brand provided in the prompt. In the ASA direction, this is reversed. The null hypothesis is that the locality of the selected/generated brand to fill in the BLANK is independent of the polarity of the attribute provided in the prompt.
We also set up our experiment where we randomly shuffle the order of choices presented to the LLMs. Thus, there is no fixed sequence where a global brand always appears first, followed by a local brand, and then a glocal brand (\Cref{fig:example}), this is also the case for attributes.
For experiment 1, we use three prompting templates in our experiments, following \citet{kamruzzaman2024banstereoset}. There is a baseline template and two paraphrases 
(see \Cref{tab:debias-persona-versions-instructions} for full templates). Our baseline prompting
template was as follows: \textit{``Based solely on the context, select the appropriate word from the options provided, without considering the order in which the options come in, only provide the appropriate word. We do not require an explanation, just a single, plain text with the answer.''}. We present our main results averaged across all three prompting templates for experiment 1.

\begin{table}[!thbp]
\begin{center}
{\small
\begin{tabular}{ |l|c c c| }
\hline

\hline
Model & Direction & $\tau$ & $p$ \\ \hline 

\multirow{2}{*}{GPT-4o} & SAI & 0.143 & \textbf{<0.001} \\
& ASA & 0.058 & \textbf{<0.001} \\ \hline
\multirow{2}{*}{Llama-3-8B} & SAI & 0.056 & \textbf{<0.001} \\
& ASA & 0.032 & \textbf{<0.001}\\ \hline
\multirow{2}{*}{Gemma-7B} & SAI & 0.089 & \textbf{<0.001} \\
& ASA & 0.001 & 0.319  \\ \hline
\multirow{2}{*}{Mistral-7B} & SAI & 0.039 & \textbf{0.009} \\
& ASA & 0.419 & \textbf{<0.001} \\ \hline
\end{tabular}}
\end{center}
\caption{\label{tab:kendall-version1} Kendall's $\tau$ test results of named variant for Experiment 1. We use a significance level of $\alpha < 0.05$ to reject the null hypothesis, in cases where the null hypothesis is rejected, we highlight these instances in bold. }
\end{table}%

\begin{table}[!thbp]
\begin{center}
{\small
\begin{tabular}{ |l|c c c| }
\hline

\hline
Model & Direction & $\tau$ & $p$ \\ \hline 

\multirow{2}{*}{GPT-4o} & SAI & 0.394 & \textbf{<0.001} \\
& ASA & 0.570 & \textbf{<0.001} \\ \hline
\multirow{2}{*}{Llama-3-8B} & SAI & 0.112 & \textbf{<0.001} \\
& ASA & 0.457 & \textbf{<0.001} \\ \hline
\multirow{2}{*}{Gemma-7B} & SAI & 0.047 & 0.1466 \\
& ASA & 0.068 & \textbf{0.004} \\ \hline
\multirow{2}{*}{Mistral-7B} & SAI & 0.137 & \textbf{<0.001}\\
& ASA & 0.516 & \textbf{<0.001} \\ \hline
\end{tabular}}
\end{center}
\caption{\label{tab:kendall-version2} Kendall's $\tau$ test results of nameless variant for Experiment 1. We use a significance level of $\alpha < 0.05$ to reject the null hypothesis, in cases where the null hypothesis is rejected, we highlight these instances in bold. }
\end{table}%





\section{Results and Discussion}

\subsection{Experiment 1 Results}

\noindent



\noindent
\Cref{tab:kendall} shows the results of the Kendall's $\tau$ test for each model and in each direction. The null hypothesis is rejected in all eight settings. 

This serves as a clear indication of a pattern of brand bias in modern LLMs. Since all $\tau$ effects are positive, LLMs are more inclined to select a negative attribute in response to a local brand. The reverse is also true, LLMs are more inclined to assume a local brand in the face of a negative attribute. In the SAI direction, we can see that GPT-4o exhibits the strongest associations (highest $\tau$ values). In the ASA direction, GPT-4o and Mistral exhibit the strongest associations.

\begin{table}[ht]
\centering
{\small
\setlength{\tabcolsep}{3.0pt}
\begin{tabular}{|c|c|c|c|c|}
\hline
\multirow{2}{*}{\textbf{Model}} & \multicolumn{2}{c|}{\textbf{\begin{tabular}{@{}c@{}}High Income\\ Countries \end{tabular}}} & \multicolumn{2}{c|}{\textbf{\begin{tabular}{@{}c@{}}Low Income\\ Countries \end{tabular}}} \\ \cline{2-5} 
                        & Luxury & Non-luxury & Luxury & Non-luxury \\ \hline
GPT-4o                  &  98.88      &   1.11   &  1.97    &   98.02              \\ \hline
Llama-3-8B             &    98.46    &     1.53       &   2.80     &  97.19           \\ \hline
Gemma-7B                &    100    &      0.0      &  16.05      &  83.94           \\ \hline
Mistral-7B              &   88.21     &   11.78         &    11.78    &  88.21           \\ \hline
\end{tabular}
}
\caption{\label{tab:category-2} Experiment 2's results, averaged across all brands. All the results are presented as the percentage (\%). }
\end{table}


We now look at the results broken down by brand type, where the $\tau$-test results are presented in \Cref{tab:statistical test for each category}. There seem to be no broad patterns in terms of which model-direction-category combinations are not statistically significant. We see that Llama-3-8B and Mistral-7B are consistently biased in the ASA direction, Gemma-7B and GPT-4o are consistently biased in the SAI direction. Otherwise, exceptions exist.

\begin{table*}[ht]
\centering
{\small
\setlength{\tabcolsep}{4.0pt}
\begin{tabular}{|c|c|c|c|c|c|c|c|c|}
\hline
\multirow{2}{*}{\textbf{Model}} & \multicolumn{2}{c|}{\textbf{Shoe}} & \multicolumn{2}{c|}{\textbf{Clothing}} & \multicolumn{2}{c|}{\textbf{Beverage}} & \multicolumn{2}{c|}{\textbf{Electronics}} \\ \cline{2-9}
                       & Global & Local & Global & Local & Global & Local & Global & Local \\ \hline
GPT-4o                 &   1.03     &  98.96     &   0.51     &    99.48   &    0.00    &    100   &   98.44     &   1.55    \\ \hline
Llama-3-8B            &   12.43     &    87.56   &   17.09     &  82.90     &   7.77     &   92.22    &   22.79     &    77.20   \\ \hline
Gemma-7B               &   100     &   0.0   &   79.27     &    20.72   & 27.97       &   72.02    &   100     &   0.0    \\ \hline
Mistral-7B             &   1.55     &    98.44   &    0.0    &  100     &   0.0     &   100    &   92.74     &    7.25   \\ \hline
\end{tabular}
}
\caption{\label{tab:detail-category-3} Brand specific results for Experiment 3. All the results are presented as the percentage (\%).}
\end{table*}

In terms of effect sizes, we find that GPT-4o tends to have the largest effect sizes. Shoes and electronics have stronger effect sizes compared to clothing and beverages. The only negative (near-zero) effect size we observe is Gemma-7B for beverages in the ASA direction.

When we use generic phrases (`global brand' and `local brand') in lieu of specific brand names~(\Cref{par:another-confound}) to control for data scarcity issues, the bias patterns still hold~(\Cref{tab:kendall-version1} and \Cref{tab:kendall-version2}). 
Effect sizes tend to be larger in the nameless subset. This means that using generic terms tends to amplify the brand biases being measured here.

In \Cref{tab:kendall-version1}, we observe that the null hypothesis can be rejected for all cases, except for Gemma in the ASA direction. In the SAI direction, GPT-4o exhibits the largest effect size, indicated by a high $\tau$ value. Conversely, in the ASA direction, Mistral demonstrates the highest effect size.

In \Cref{tab:kendall-version2}, it is evident that the null hypothesis can be rejected for all cases, with the sole exception of Gemma in the SAI direction. Additionally, this nameless variant demonstrates a higher effect size compared to the named variant. In both the SAI and ASA directions, GPT-4o exhibits the highest effect size, indicated by a high $\tau$ value.

In Experiment 1, we excluded certain examples due to invalid responses where LLMs did not select from our list of three choices. For more details, see \Cref{app:interesting-result}.


\subsection{Experiment 2 Results}


\noindent
\Cref{tab:category-2} shows the socio-economic bias results averaged across all brands.
We observe consistent results across all models, where each suggests luxury brands as gifts for individuals from high-income (high GDP per capita) countries. Conversely, all models recommend non-luxury brands for people from low-income countries. Despite allowing the models the flexibility to suggest the same brands for both high and low-income individuals, discrepancies still arise. This suggests an inherent bias toward high and low-income countries. For results of each brand type, see \Cref{tab:detail-category-2} in \Cref{app:detailed-category-2-3}.

\subsection{Experiment 3 Results}

\begin{table}[ht]
\centering
{\small
\setlength{\tabcolsep}{3.0pt}
\begin{tabular}{|c|c|c|}
\hline
\textbf{Model} & \textbf{Global brand (\%)} & \textbf{Local brand (\%)} \\ \hline
GPT-4o & 25.00 & 75.00 \\ \hline
Llama-3-8B & 15.04 & 84.95 \\ \hline
Gemma-7B & 76.81 & 23.18 \\ \hline
Mistral-7B & 23.57 & 76.42 \\ \hline
\end{tabular}
}
\caption{\label{tab:category-3} Experiment 3's results, averaged across all brands. }
\end{table}


\noindent

In \Cref{tab:detail-category-3}, we represent our experiment 3 results for each brand type. From \Cref{tab:detail-category-3}, it is evident that most models prefer local brands when the domestic country is specified. GPT-4o generally favors local brands across various brand types, with the exception of electronics, where it prefers global brands. Lllama-3 consistently favors local brands in all categories. In contrast, Gemma tends to favor global brands, except in the beverage category where it shows a preference for local brands. Similar to GPT-4o, Mistral primarily favors local brands.

We represent our experiment 3's results, averaged across all brands in \Cref{tab:category-3}. We observe that all models except Gemma suggest local brands when the country name is included in the prompt. This indicates the country of origin effect, where models tend to favor brands from their own countries over global brands. However, Gemma shows a preference for global brands regardless. 

\section{Conclusion}

This study reveals that LLMs exhibit brand biases, particularly favoring global brands over local ones, which could affect consumer behavior and brand perception. This could exacerbate challenges for local brands competing in a global market, inadvertently favoring already dominant global brands and potentially stifling competition and innovation. Additionally, socio-economic biases affect LLM recommendations between luxury and non-luxury brands, correlating with a country's wealth. Finally, we found global brand bias may reverse due to country-of-origin effects, revealing potential complications in broad LLM behavior generalizations.


\section{Limitations}
Several factors in our experiments may restrict the generalizability of our results and conclusions. Although we selected four of the most prevalent and advanced LLMs currently available, our study was not exhaustive. Numerous LLM variants exist today and more will likely emerge in the future. Also, we only consider four types of brands, but there are numerous types of brands available. In Experiment 1, we collected actual brand names exclusively from 15 countries. Although these were chosen to represent a diverse array of brands, they might not cover all brand categories or geographic regions comprehensively. In Experiment 2, we considered only 30 countries with higher GDP and 30 with lower GDP, which may not fully represent global diversity. Additionally, we solely used socio-economic conditions (GDP per capita) to assess the impact, yet LLMs may also harbor biases related to other social factors such as skin color, gender, and occupation. It would be beneficial to explore these variables to understand how LLMs perform under varying conditions. Our experiments were conducted exclusively in English. This limitation means the behavior of LLMs, which are capable of operating in multiple languages, might differ in non-English contexts.

\section*{Acknowledgements}
This project was fully supported by the University of South Florida. We thank the reviewers for their valuable feedback in this paper.

\bibliography{anthology,custom}

\appendix

\section{Collection of Attributes and Stimuli}
\label{app:collection-detail}
\paragraph{Attributes. } We employed a human-LLM collaboration to gather attributes. We instructed ChatGPT-4 to provide 40 positive, negative, and neutral attributes for each brand using the prompt: `Give me 40 [positive/negative/neutral] attributes to describe [brand type]'. In this prompt, we substituted [brand type] with the specific brand. After receiving 40 positive, 40 negative, and 40 neutral attributes for each brand, we applied human filtration to determine the appropriateness of these attributes in our sentences and context. We had four human annotators; an attribute was retained if it was deemed suitable by at least three annotators. Ultimately, we selected nine attributes from each category—positive, negative, and neutral.





\paragraph{Stimuli. } 
We encounter challenges in collecting brand names, especially those of local brands. We manually gather both local and global brand names from 15 countries.\footnote{Due to the challenges associated with collecting local brand names, and our requirement to use at least three local brand names from each country, we restrict our experiments to 15 countries.} We collect 3 global and 3 local brand names from each county. We select 5 countries with high GDP per capita, 5 with average GDP per capita, and 5 with low GDP per capita (data available).\footnote{We use GDP per capita values as reported by the International Monetary Fund (IMF) (as of May 2024).}\textsuperscript{,}\footnote{\url{https://www.imf.org/external/datamapper/NGDPDPC@WEO/OEMDC/ADVEC/WEOWORLD}} A brand is classified as local if it lacks an in-person showroom outside its country of origin. To ensure this, we verify the absence of any in-person showrooms by consulting the brand's official website to view all their store locations or by searching for the brand's showrooms outside the region if the website is not well-established. We consider a brand as a global brand if it operates and is available in multiple countries across various continents. This does not imply that it must be present in every country; however, it should maintain a significant presence in major markets worldwide. We also cross-verify this collection by two human annotators.\footnote{We strive to ensure that local brands do not have any in-person showrooms outside their country of origin. However, due to the lack of a comprehensive website and availability of data, we may overlook existing showrooms beyond that region. }

\section{Model Details}
\label{app:model-detail}
We use four major language models for assessing our task: 1) The GPT-4o using checkpoint on the OpenAI API; 2) Llama-3-8B via the meta-llama
/Meta-Llama-3-8B-Instruct checkpoint on Huggingface; 3) Mistral-7B via the TheBloke/Mistral-7B-Instruct-v0.1-GGUF checkpoint on Huggingface, 4) Google's open source model Gemma-7B using Ollama framework.




\section{Detailed Prompting Templates}
\label{app:prompt}

\begin{table*}
\centering
\begin{tabular}{ p{0.20\textwidth}|p{0.70\textwidth} }
\hline
\textbf{Prompt Type} & \textbf{Prompt} \\
\hline
Baseline Prompt &  Based solely on the context, select the appropriate word from the options provided, without considering the order in which the options come in, only provide the appropriate word. We do not require an explanation, just a single, plain text with the answer. \\
\hline
Rephrase prompt 1 & Choose the appropriate word from the options given based only on the context. Provide just the selected word, without considering the order of the options, and no explanation is needed. \\
\hline
Rephrase prompt 2 & From the given context, choose the appropriate word from the provided options without considering their order, and provide only the word as a plain text answer without any explanation.\\
\hline
\end{tabular}
\caption{Baseline prompt and two other prompts by rephrasing the baseline prompt. }
\label{tab:debias-persona-versions-instructions}
\end{table*}

\begin{table*}[ht]
\centering
{\small
\setlength{\tabcolsep}{2.5pt}
\begin{tabular}{|c|c|c|c|c|c|c|c|c|c|c|c|c|c|c|c|c|c|}
\hline
\multirow{3}{*}{\textbf{Model}} & \multicolumn{8}{c|}{\textbf{High-income countries}} & \multicolumn{8}{c|}{\textbf{Low-income countries}} \\ \cline{2-17}
                       & \multicolumn{2}{c|}{Shoe} & \multicolumn{2}{c|}{Clothing} & \multicolumn{2}{c|}{Beverage} & \multicolumn{2}{c|}{Electronics} & \multicolumn{2}{c|}{Shoe} & \multicolumn{2}{c|}{Clothing} & \multicolumn{2}{c|}{Beverage} & \multicolumn{2}{c|}{Electronics} \\ \cline{2-17}
                       & L & NL & L & NL  & L & NL  & L & NL  & L & NL  & L & NL  & L & NL  & L & NL  \\ \hline
GPT-4o                 &    98.54    &    1.45         &   98.43     &   1.56          &    99.79    &      0.20       &   98.75     &     1.24        &    1.04    &      98.95       &   2.39     &     97.60        &   3.85     &     96.14        &    0.62    &     99.37       \\ \hline
Llama-3-8B            &    99.06   &     0.93        &    97.71    &   2.28          &    98.75    &    1.24         &    98.33    &     1.66        &    1.97    &   98.02          &    3.12    &      96.87       &   4.05     &      95.94       &   2.08     &      97.91       \\ \hline
Gemma-7B               &    100    &    0.0         &    100    &   0.0          &   100     &    0.0         &    100    &  0.0  &   8.94     &   91.05          &  14.67      &      85.33       &    36.62    &      63.37       &    3.95    &     96.04              \\ \hline
Mistral-7B             &    86.57    &       13.42      &    86.99    &  13.00           &   93.96     &       6.03     &    85.04    &     14.95        &    13.42    &      86.57       &    13.00    &    87.00         & 6.03       &     93.96        &   14.67     &    85.32         \\ \hline
\end{tabular}
}
\caption{\label{tab:detail-category-2} Brand specific results for Experiment 2. Here L stands for Luxury, NL stands for Non-luxury. All the results are presented as the percentage (\%). }
\end{table*}



\section{Detailed Result Section for Experiment 2}
\label{app:detailed-category-2-3}

In \Cref{tab:detail-category-2}, we represent our experiment 2 results for each brand type. From \Cref{tab:detail-category-2}, it is evident that all models exhibit a preference for luxury brands for individuals from high-income countries. Gemma demonstrates a pronounced bias, with 100\% of its suggestions for high-income individuals being luxury brands. However, for individuals from low-income countries, while Gemma still assigns luxury brands, the percentage is significantly lower.


\begin{table*}
\centering
{\small
\begin{tabular}{|c|c|c|c|c|c|c|}
\hline
Models & \begin{tabular}{@{}c@{}}Non-Option \\Span\end{tabular} & \begin{tabular}{@{}c@{}} No \\ Response\end{tabular} &  \begin{tabular}{@{}c@{}}Out-of-Context \\ Responses\end{tabular} & \begin{tabular}{@{}c@{}}Approximate \\ Match\end{tabular} & Total\\
\hline
GPT-4o & 3 & 0 & 0 & 32 & 35\\
\hline
Llama-3-8B & 50 & 6 & 4 & 34 & 94\\
\hline
Gemma-7B & 34 & 138 & 0 & 21 & 193\\
\hline
Mistral-7B & 214 & 466 & 21 & 123 & 824\\
\hline
\end{tabular}
}
\caption{Number of Invalid Responses for Experiment 1}
\label{tab:invalid-response}
\end{table*}

\section{Invalid Results Categorized}
\label{app:interesting-result}


Based on the pattern of broad invalid responses, we have categorized the responses into four distinct groups. Category 1, labeled ``Non-Option Span'', comprises responses that originate from the context sentence but not from the option list. Category 2, named ``No Response'', includes cases where the sentence remains incomplete, either because the response is null or assistance cannot be provided. Category 3, referred to as ``Out-of-Context Responses'', encompasses responses that fall outside both the context sentence and the option list. Category 4, titled ``Approximate Match'', consists of responses that are closely related to the provided stimulus or attribute but do not replicate it exactly, often due to minor inaccuracies or variations in wording. Table~\ref{tab:invalid-response} shows the exact number of invalid responses in each category across all models. In many instances, invalid responses included elements from the provided prompt sentence but not from the option list which is Category 1 ``Non-Option Span''. Another frequent scenario was responses that were closely related to the provided stimulus or attribute but were not exact replicas (Category 4: ``Approximate Match''). The most common type of response in the ``No Response'' category was simply ``blank'', used as a placeholder in the sentence the LLM was required to complete.

\section{Lists of Attributes and Stimuli}
\label{app:attribute-list}

\begin{table*} [!thbp]
\begin{center}
{
\small
\setlength{\tabcolsep}{2.5pt}
\begin{tabular}{|c|c|c|| c|c|c|  }
\hline
\multicolumn{6}{ |c| }{\textbf{SAI Direction}} \\

\hline
 & \multicolumn{2}{ c|| }{\textbf{{Stimulus}}} & \multicolumn{3} { c| } {\textbf{{Attributes}}} \\
\hline
Brand Type & Positive & Negative & Positive & Negative & Neutral \\
\hline
Shoe & Nike, Skechers, ... & Jmofs, Sikhar, ...  & \begin{tabular}{@{}c@{}}fashionable, \\comfortable,...\end{tabular} & \begin{tabular}{@{}c@{}}unfashionable, \\ uncomfortable,...\end{tabular} & functional, average, ... \\ \hline
Clothing & \begin{tabular}{@{}c@{}}Calvin Klein, \\ Levis’s, ...\end{tabular} & Jaad, Bigi, ... & \begin{tabular}{@{}c@{}}well-designed, \\ stylish,...\end{tabular} & \begin{tabular}{@{}c@{}}poorly designed, \\ outdated,...\end{tabular} &  alright, okay, ... \\ \hline
Beverage & \begin{tabular}{@{}c@{}}PepsiCo, \\ Coca–Cola, ...\end{tabular} & Flauder, Cristal, ... & refreshing, smooth,... & \begin{tabular}{@{}c@{}}flat, \\ rough,...\end{tabular} & reasonable, plain,...\\ \hline
Electronics & Apple, Intel, ... & \begin{tabular}{@{}c@{}}Centelsa, \\ Electro Max, ...\end{tabular} & durable, reliable, ... & fragile, unreliable,... & subdued, retro, ... \\
\hline 

\end{tabular}
}
\end{center}
\caption{\label{tab:stimulus-attribute list-sai} List of Stimulus and Attributes in SAI direction. These are not actual representations. For the sake of our writing, we consider this way.   }
\end{table*}

\begin{table*} [!thbp]
\begin{center}
{
\small
\setlength{\tabcolsep}{2.5pt}
\begin{tabular}{|c|c|c|| c|c|c|  }
\hline
\multicolumn{6}{ |c| }{\textbf{ASA Direction}} \\

\hline
 & \multicolumn{2}{ c|| }{\textbf{Attributes}} & \multicolumn{3} { c| } {\textbf{{Stimulus}}} \\
\hline
Brand Type & Positive & Negative & Positive & Negative & Neutral \\
\hline
Shoe & \begin{tabular}{@{}c@{}}fashionable, \\comfortable,...\end{tabular} & \begin{tabular}{@{}c@{}}unfashionable, \\ uncomfortable,...\end{tabular} & Nike, Skechers, ... & Jmofs, Sikhar, ... & `a glocal brand' \\ \hline
Clothing & \begin{tabular}{@{}c@{}}well-designed, \\ stylish, ...\end{tabular} & \begin{tabular}{@{}c@{}}poorly designed, \\ outdated, ...\end{tabular} & \begin{tabular}{@{}c@{}}Calvin Klein, \\ Levis’s, ...\end{tabular} & Jaad, Bigi, ... &  `a glocal brand' \\ \hline
Beverage & refreshing, smooth,... & \begin{tabular}{@{}c@{}}flat, \\ rough,...\end{tabular} & \begin{tabular}{@{}c@{}}PepsiCo, \\ Coca–Cola, ...\end{tabular} & Flauder, Cristal, ... & `a glocal brand' \\ \hline
Electronics & durable, reliable, ... & fragile, unreliable,... & Apple, Intel, ... & \begin{tabular}{@{}c@{}}Centelsa, \\ Electro Max, ...\end{tabular} & `a glocal brand' \\
\hline 

\end{tabular}
}
\end{center}
\caption{\label{tab:stimulus-attribute list-asa} List of Stimulus and Attributes in ASA direction. These are not actual representations. For the sake of our writing, we consider this way.   }
\end{table*}

\end{document}